\documentclass{article} 
\usepackage{nips12submit_e,times}
\usepackage{subfigure} 
\usepackage{graphicx}
\usepackage{amsmath}
\usepackage{color}
\usepackage{amssymb}
\usepackage{algorithm}
\usepackage{algorithmic}
\usepackage{multirow}

\newcommand{\vect}[1]{\boldsymbol{#1}}

\usepackage[turnoff]{notes}

\title{Structuring Relevant Feature Sets with Multiple Model Learning }

\author{
Jun Wang \\
Department of Computer Science\\ 
University of Geneva\\ Switzerland\\
\texttt{Jun.Wang@unige.ch} 
\And Alexandros Kalousis \\
 Department of Business Informatics\\
 University of Applied Sciences \\Western Switzerland\\
\texttt{Alexandros.Kalousis@hesge.ch} }

\newcommand{\ifg}{NR}
\newcommand{\cfs}{FSSFS}
\newcommand{\egd}{EGDM}
\newcommand{\mdsvm}{MD-SVM}

\nipsfinalcopy 

\begin{document}

\maketitle

\begin{abstract}
Feature selection is one of the most prominent learning tasks, especially in high-dimensional datasets in which 
the goal is to understand the mechanisms that underly the learning dataset. However most of them typically 
deliver just a flat set of relevant features and provide no further information on what kind of structures, e.g.
feature groupings, might underly the set of relevant features. In this paper we propose a new learning paradigm
in which our goal is to uncover the structures that underly the set of relevant features for a given learning problem.
We uncover two types of features sets, non-replaceable features that contain important information about the target
variable and cannot be replaced by other features, and functionally similar features sets that can be used interchangeably 
in learned models, given the presence of the non-replaceable features, with no change in the predictive performance. To 
do so we propose a new learning algorithm that learns a number of disjoint models using a model disjointness regularization
constraint together with a constraint on the predictive agreement of the disjoint models. We explore the behavior of our
approach on a number of high-dimensional datasets, and show that, as expected by their construction, these satisfy a number 
of properties. Namely, model disjointness, a high predictive agreement, and a similar predictive performance to models 
learned on the full set of relevant features. The ability to structure the set of relevant features in such a manner 
can become a valuable tool in different applications of scientific knowledge discovery. 
\end{abstract}

\section{Introduction}
\label{introduction}
Feature selection\cite{guyon2003introduction} is one of the most often performed tasks in supervised learning problems, especially when the goal is to gain an understanding 
of the mechanisms that underly some, often high dimensional, learning dataset. Such analysis scenarios are typical in scientific knowledge discovery,
with biology providing ample examples. However existing feature selection and classification algorithms provide at most a flat list of relevant 
features, with no further information on the internal structure of that feature set. Nevertheless it is now a well known fact that within a set of 
relevant features for a given problem there can be a number of different models defined over different feature subsets which nevertheless 
are of high predictive power~\cite{Ein-DorEtAl2005,boutros2009prognostic}. A typical such scenario appears in problems with high levels of feature redundancy.

In this paper we want to go one step further and uncover the structure underlying the set of relevant features for a given learning problem. 
We will do so by learning within this feature set as many as possible structurally dissimilar models, i.e. models defined over different feature 
subsets of the set of relevant features. Nevertheless we will constrain these models to have a very similar predictive behavior in terms of the 
predictions they make, and a very high predictive power, similar to that which a model learned on the full set of relevant features would achieve. 
By learning different models which have these properties we expect that highly complimentary feature sets with respect to the target variable 
are placed together in the individual models, while highly redundant feature sets will end up in different models. We will further structure
the features used by these basis models in two basic feature sets. One feature set will be the non-replaceable features, i.e. features that
will be systematically present within all the basis models learned. This set of features is critical for the accurate description of the 
target variable and their removal from a basis model would result to a loss of predictive power. In addition to that set 
we will have the set of the compliment feature sets of the non-replaceable feature set defined over the different basis models. These 
compliment feature sets have a similar information content with respect to the target variable given the set of non-replaceable 
features; each one of them can be used instead of another without any significant change in the predictive behavior or
performance. 
The availability of such a structure can provide us with a much better insight to the 
learning problem that is studied. This has the potential to be a game-changing technique 
especially in problems in which understanding the mechanisms that underly the learning problem 
is what drives the data analysis process.

\note[Removed]{
Since the definition of EGD models are strict, one might question whether there is the existence of EGD models. However, with the 
abundant of high dimensional application, we do believe the existence of EGD models. For instance, in microarray classification problem, 
the existence of multiple predictive sets is a well known fact. Learning this kind of feature structure based on EGD models can be potentially 
applied into many applications. For instance, in biological problem identifying the feature structure of a highly predictive model could provide 
valuable scientific insight into the learning problem, since the different feature groups in CFSFGs set might correspond to redundant mechanisms 
underpinning the study problem. Furthermore, the knowlodge of CFSFGs set could be also extremely valuable for many application domain whose 
features are cost-sensitive. For instance, in the designing of chemistry compound the cost of elemental materials are often very different. 
With the knowledge of feature structure, we can reduce cost by simply using the cheapest materials from CFSFGs set.  }

To the best of our knowledge there exist no learning approaches that are able to uncover structures within a set of relevant feature such
as the ones just described. Standard feature selection and classification
algorithms as already mentioned return only a flat set of relevant feature sets, often
accompanied by their relative importance in terms of some ranking score.  
A rather simplistic approach that is often used to structure the 
set of features relies on the use of pairwise feature redundancies estimated through some feature similarity measure. Most often these 
approaches take the form of feature clustering which uses as a feature similarity measure some measure of feature correlation, 
placing like that in the same cluster features with a high degree of pairwise 
redundancy. Nevertheless, the target of this rather different feature structure is often to provide background knowledge for regularizing further the model fitting~\cite{tibshirani2005sparsity,ZhongK11,Loscalzo2009}.

\note[Removed]{Standard feature selection algorithm can present a unique feature set by distinguishing the relevant and irrelevant 
features. It could know the importance of each features with respect to the learning problem by a ranking score. However, it has no knowledge 
about the internal structure of selected features. An alternative heuristic approach to learn the structure of features is through the clustering 
algorithm. Based on a feature similarity, the features can be divided into several groups. However, this kinds of feature structure does not 
consider the predictive power of feature groups and thus could not provide much insight about the learning problem. }

A central component in structuring the set of relevant features in the manner described above is to come up with a way to learn 
as many as possible equally good but dissimilar models. In this paper we will present a novel multiple model learning algorithm that does
exactly that. We will take standard objective functions such as the ones used in learning simple linear models and use them to simultaneously 
learn a number of dissimilar models by coupling them with a novel disjointness regularization which will force the learned models to use 
different discriminative features. In order to guarantee that all models will be of roughly equal predictive power and equivalent predictive
behavior we will also regularize them in a manner that will force them to produce very similar predictions.
We will will demonstrate the utility of the novel learning task on a number of high-dimensional microarray classification problems.

%

The rest of the paper is organized as follows. In section~\ref{sec:FSFGL} we will introduce a number of necessary definitions which we will 
use to describe in section~\ref{sec:MDML} our approach to learning multiple models and the respective optimization problem, in 
section~\ref{sec:opt} we show how to solve the latter. In section~\ref{sec:exp} we present our experiments, and we conclude in section~\ref{sec:con}.

\section{Learning the Structure of a Set of Predictive Features}
\label{sec:FSFGL}
The new learning task that we want to define and address is to uncover the structure underlying a set of 
features for a given learning problem. For simplicity in this paper we will learn the structure of a {\em relevant} feature set. Nevertheless, our approach can be used to uncover the structure of a feature set with irrelevant features. However, in this case a sparse learning algorithm with embedding feature selection strategy should be used to remove the irrelevant features.

The type of structure that we wish to uncover is the separation of the relevant 
feature set to a set of non-replaceable features and a set of functionally similar feature sets given the non-replaceable set of features.
We assume that the original set of relevant features will 
be either given or we will establish it with the help of some feature selection algorithm. In the experimental part of this paper
we will use SVM with elastic net regularization (EN-SVM)~\cite{wang2006doubly} and retain only features with a non-zero coefficient to determine the initial 
set of relevant features.


We denote by $\mathbf X$ the $n \times d$ matrix  of learning instances for the given relevant feature set $S=(s_1,\dots,s_d)$. 
The  $i$-th row $\mathbf X$ is the $\mathbf x_i^T \in \mathbb{R}^d$ instance; the vector 
$\mathbf y=(y_1,\ldots, y_n)^T$, $y_i \in \{-1,1\}$ is the vector of class labels, for simplicity we will only consider binary 
classification. The extension to multiclass 
classification will be discussed later. To structure the set $S$ we will rely on models that will be learned with the help
of some linear algorithm. 
We will use a linear SVM with $\ell^2_2$ regularization~\cite{cortes1995support}, and we will 
denote a  model learned on $S$ by  $\vect w_S = (w_1, \dots, w_d)$, by $e(\vect w_S)$ its predictive error, and by $\vect w_S(\mathbf x)$
its prediction for the $\mathbf x$ instance, when it is clear from the context we will omit the subscript $S$ which indicates
the feature set on which the model was learned. Additionally we will make use of the concept of predictive agreement of
$m$ models $\vect w_1,\cdots,\vect w_m$ which we define as $PA(\vect w_1,\cdots,\vect w_m,\mathbf x)=\frac{\sum_{i,j,i\neq j}P(\vect w_i(\mathbf x)==\vect w_j(\mathbf x))}{m(m-1)}$, where $P(\vect w_i(\mathbf x)==\vect w_j(\mathbf x))$
is the probability of that given some instance $\mathbf x$ the two
models $\vect w_i$ and $\vect w_j$ produce the same prediction. 
Empirically we evaluate it on a dataset $\mathbf {X_t}$ of $l$ instances by $P(\vect {w_i}, \vect {w_j},\mathbf {X_t})=\frac{\sum^{k=l}_{k=1}\delta(\vect w_i(\mathbf {x_k}),\vect w_j(\mathbf {x_k})}{l}$, where $\delta(a,b)=1$ if $a=b$, and $0$ otherwise.
We will define the dissimilarity of two feature sets $A_i$ and $A_j$ by $D(A_i, A_j)=1-\frac{|A_i \cap A_j|}{A_i \cup A_j},D(A_i, A_j) \in [0,1]$ and by
$Dis(A_1,\cdots, A_m)=\frac{\sum_{i,j,i\neq j}D(A_i, A_j)}{m(m-1)}$ the dissimilarity for $m$ feature sets $A_1,\cdots,A_m$.
Finally we will call these $m$ feature sets {\em non-trivially dissimilar} if for any pair feature sets $1-D(A_i, A_j) < \theta$, where $\theta$ is a small positive value, e.g. $0.6$.
\note[Alexandros]{Maybe we should define a $\theta$-non-trivial dissimilarity level, i.e. make it a user set parameter. 
I think that this $\theta$ parameter will play a role that is similar to the sparsity level parameter used in sparse 
learners. Here instead of the sparsity it will control the dissimilarity, and it can even be a parameter to tune?}
\note[Theta]{
Finally we will call two feature sets 
$A_i \subseteq S$ and $A_j \subseteq S $ {\em $\theta$-dissimilar} if some bounded measure of their 
overlap has value smaller than $\theta$, where $\theta$ some problem-dependent value of feature 
overlap, e.g. $o(A_i, A_j)=\frac{|A_i \cap A_j|}{A_i \cup A_j} < \theta, o(A_i, A_j) \in [0,1]$. We 
will call them non-trivially dissimilar if $\theta$ is a small value
}

In the following section we will provide the 
definitions of the main concepts that we will be using in the problem of uncovering the structure of a set of 
predictive features.


\subsection{Key Definitions}
\label{sec:PMDM}
We will start by giving the definition of equally good and dissimilar models, \egd, with the help of predictive error
and predictive model agreement. We will subsequently use the definition of \egd\ to define the concept of non-replaceable
feature set, \ifg, followed by the concept of functionally-similar set of feature sets conditioned on some feature set, \cfs . 
The \ifg\ and the set of \cfs  correspond to the structure of the feature set that we wish to uncover. 


\newtheorem{definition}{Definition}
\begin{definition}
\label{EGD}
Given  $S$ a set of relevant features, a set of $m$ models $\vect w_{A_1}, \cdots, \vect w_{A_m}$, learned over non-trivially dissimilar 
feature subsets $A_i$ of $S$, $A_i \subseteq S$, we will call the set of $m$ models a set of Equally Good and Dissimilar Models, \egd,
if for every model we have $|e(\vect w_{A_i}) - e(\vect w_S)| < \epsilon_1 $, 
and for all models we have $1-PA(\vect w_{A_1},\cdots,\vect w_{A_m},\mathbf x) < \epsilon_2 $,
where $\epsilon_1$ and $\epsilon_2$ are some small positive values. 
\end{definition}
In other words a set of $m$ models is an \egd\ if the different models that belong to it are defined over non-trivially dissimilar 
feature subsets, have a predictive error that is almost identical to that of the model learned on the full set of relevant features $S$,
and a very high predictive agreement. From the definition it is clear that the \egd\ models are models of high predictive performance, 
almost the same as that of the full feature set, which use different discriminative features. Note that not all learning problems can 
have \egd\ models, but only those which have high, structurally, redundant information with respect to the target variable. 

\note[Theta]{
\begin{definition}
\label{EGD}
Given  $S$ a set of relevant features, a set of $m$ models $\vect w_{A_1}, \cdots, \vect w_{A_m}$, 
learned over feature subsets $A_i$ of $S$, $A_i \subseteq S$, we will call the set of $m$ models 
a set of Equally Good $\theta$-Dissimilar models, 
$\theta$-\egd, if every pair of feature sets, $A_i, A_j$, is $\theta$-dissimilar, for
every model we have $|e(\vect w_{A_i}) - e(\vect w_S)| < \epsilon $, 
and for every pair of models we have $1-P(\vect w_i(\mathbf x)==\vect w_j(\mathbf x)) < \epsilon $,
where $\epsilon$ some very small positive value, 
\end{definition}
}

\begin{definition}
\label{IFG}
A feature set $I\subseteq S$ is non-replaceable, \ifg,  if $I=\cap_i A_i$, i.e. if it
is the intersection of the feature sets used in the \egd\ models.
\end{definition}
The features of \ifg\ are not necessarily the most predictive ones. However when we want to maximize the 
predictive performance their information contribution cannot be brought in by any other subset of $S$. Note 
that in order to establish the \ifg\ of a feature set $S$ we need to discover all \egd\ models. This is an NP-hard 
problem. We will approximately learn \ifg\ by learning as many as possible \egd\ models. 

\begin{definition}
\label{FS}
We will call a set of $m$ different feature sets $F_1, \cdots, F_m$, $\bigcap^m_{i=1} F_i = \emptyset$, 
a functionally similar set of feature sets conditioned on a given feature set $C$, and denote it by \cfs, 
if the set of models $\vect w_{F_1 \cup C}, \cdots, \vect w_{F_m \cup C}$ is an \egd\ set.
\end{definition}
The information content that these $m$ feature sets deliver with respect 
to the target variable is equivalent. Learning in the presence of the $C$ feature set
any feature set of \cfs\ can be replaced by some other feature set of \cfs\ with almost 
no change in predictive behavior. \cfs\ provides us with a structure of the feature set 
in terms of sets of feature sets of similar information content with respect to the target 
variable. This together with the \ifg\ set fully structures the feature set $S$ in terms of
non-replaceable features as well as sets of features that bring the same information content.
The availability of such a structure can provide
us with a much more complete picture than that a flat list of relevant features can provide.
This is especially crucial in data analysis scenarios in which gaining an
insight to, and understanding the mechanisms that underly and produced the learning dataset is what
drives data analysis.

In the following sections we will present an algorithm which learns a set of equivalently good and dissimilar 
models satisfying the definition~\ref{EGD}. To the best of our knowledge it the first algorithm of that kind.

\note[Removed]{In order to learn a set of \egd\ that satisfy definition \ref{EGD} we need a learning algorithm which 
produces models with the following three properties. First, the learned models should use different 
discriminative features, second

achieve the conditions stated in definition \ref{EGD}, we need to learn models that satisfy the following three property. 
Property (1) The learned models should use different discriminative features. Property (2) The learned models should achieve 
(almost) the same predictive performance as the model learned with all features. Property (3) The prediction agreement between 
models should be very high. Currently, there is no learning approach learns model like that.
In the following subsections, we will show how to learn multiple models that achieves Property (1)-(3).  }

\subsection{Multiple \egd\ Learning}
\label{sec:MDML}

At the core of our approach we have the learning of multiple equivalent good and distinct learning models. 
Since we will be given a set of relevant features over which we will learn our multiple models here we do 
not need to use any sparsity constrain on the features because we know apriori that all of them are relevant
and we want to have models of predictive performance that is similar to what we can achieve if we use the 
full feature set. We can make use of standard cost functions found in non-sparse linear classification algorithms such as linear 
SVMs and logistic regression with $\ell_2^2$ penalty. However we will not learn these models independently of
one another. We will learn them collectively and force them to be dissimilar by using different discriminative
features. One cannot help but think a symmetry with multi-task learning~\cite{caruana1997multitask}. In the latter we learn similar models
over different datasets, here we learn dissimilar models over the same dataset.

We will control the model dissimilarity through the introduction of a disjointness regularization term.
More precisely, we define the disjointness regularization term for $m$ models,  $\vect {w_1},\cdots,\vect {w_m}$, as:
\begin{eqnarray}
\label{disjoint-reg}
\Omega(\vect {w_1},\cdots,\vect {w_m})=\sum_{ij, i\neq j}\left|\vect{w_i}\right|^T\left|\vect{w_j}\right|
=\sum_{l}\sum_{ij, i\neq j}\left|w_{il}w_{jl}\right|\\\nonumber
\end{eqnarray}
This regularization is motivate by a simple observation. If two models are totally disjoint, 
their entry-wise product will be the $\vect{0}$ vector. Based on this observation we regularize the sum of 
the sparsity-inducing $\ell_1$ norm on the element-wise product of all model pairs.  Minimizing this sum will 
push most of the entries of the element-wise model products to be $0$; as a result different 
models will select different features. Note that there is no disjointness regularization on the intercepts of the models.

\note[Removed]{
To achieve the Principle (2), we will rely on a standard learning algorithm. Since we will be learning EGD models in a 
predictive feature set selected by a feature selection algorithm, principally all the features are useful for the learning problem. As a result, 
we will prefer to learning model by a non-sparse linear classification algorithm, such as linear SVM and logistic regression with $\ell_2^2$ penalty. 
}

To learn the different models we will use the linear SVM objective function, i.e. we will minimize the trade-off of the 
margin, i.e. the $\ell_2^2$ norm of the normal vectors, and the hinge loss error. In addition we want the different models
to produce similar predictions, to do so we will add an additional term which will penalize the prediction dissagreement.
We will only constrain the models to predict the same class label, an even stronger constraint would force them to produce
exactly the same output value. The final optimization problem will be created from the combination of the above elements, 
i.e. the SVM objective function, the prediction agreement constraints, and the disjointness regularization term, and will 
be:
\begin{eqnarray}
\label{multi-1norm-svm}
\min_{b_1,\cdots,b_m,\vect{w_1},\cdots,\vect{w_m},\vect \epsilon, \vect s}&&\sum_{i=1}^m\sum_{k=1}^{n}\epsilon_{ik}+\lambda_1\sum_{i,j, i \neq j}^m\sum_{k=1}^{n}s_{ijk}\\\nonumber
&&+\frac{\lambda_2}{2} \sum_{ij, i\neq j}\left|\vect{w_i}\right|^T\left|\vect{w_j}\right| +\lambda_3 \sum_{i=1}^m\left\| \vect{w_i} \right\|^2_2\\\nonumber
s.t.&&  y_k(b_i+\vect{w_i^T x_k}) \geq 1- \epsilon_{ik}, \epsilon_{ik} \geq 0, \forall {i,k}\\\nonumber
&&  (b_i+\vect{w_i^T x_k}) *  (b_j+\vect{w_j^T x_k}) \geq -s_{ijk}, s_{ijk} \geq 0, \forall {i,j,k}\\\nonumber
\end{eqnarray}
This optimization problem learns $m$ dissimilar models. We maximize prediction agreement by penalizing the 
prediction disagreement, defined in the second constraint, and we control its importance through the $\lambda_1$ 
parameter. We control the importance of the model dissimilarity through the trade-off defined by the $\lambda_2$ 
and $\lambda_3$ parameters over the model dissimilarity and the $\ell_2^2$ model norms. We should emphasize that 
the use of the $\ell_2^2$ model norms has an additional advantage. In combination with the model dissimilarity term, 
as elastic net regularization~\cite{zou2005regularization}, it forces highly correlated features either to be included all together in the same model or excluded all together. The result 
is that like that we do not discover dissimilar models which would be created simply because highly correlated
features are placed in different models, producing models with very similar predictive performance and on the same time 
high model dissimilarity. It is this property that guarantees the discovery of valuable feature groupings.
We will call the optimization problem given in~(\ref{multi-1norm-svm}) Multiple Dissimilar SVMs, \mdsvm s. 

\note[Removed]{
In the absence of any prior knowledge all individual models will have the same value for all parameters. Nevertheless it is possible to have 
models with different levels of sparsity and dissimilarity to each other in which cases  different parameters should be adapted accordingly. 
We will call the resulting method Multiple Dissimilar SVMs (MDSVMs). }

Since all models learned by \mdsvm s should achieve the same target, i.e. minimize the two hinge loss errors and the $\ell_2^2$ regularization, 
the result is that they compete against each other to select the most useful features. Features that are most important in predicting the
target variable and cannot be replaced by other features will be present in all models giving rise to the \ifg\ feature set. Feature sets that
are highly complimentary to the \ifg\ and structurally redundant between them will end up in different models producing the different feature
sets of the \cfs\ conditioned on the \ifg\ set.

\note[Removed]{
are hopefully selected together. In such a way, the individual model learned by MDSVMs can give more useful interpretation 
of the learning problem, comparing with the single model learned by 1-norm SVM, which potentially involves the most useful features from different groups. 
It is worthy to note that we expect the predictive performance of single model learned by 1-norm SVM could be slightly better than that of models learned 
by MDSVMs. The reason is the single model involves the most useful features, in contrast, the models learned by MDSVMs select different correlated feature groups.
}

Before describing in the next section how we will solve the optimization problem of \mdsvm s we will briefly discuss some related work on dissimilar 
model learning.  In fact there has been very limited work on learning dissimilar models and this in a quite different context. \cite{zhou2011hierarchical,hwanglearning} 
learn dissimilar models for the tree-structured multi-class classification problem. The different models are learned over different sub-classification 
tasks of some given multi-class classification problem. 
Their motivation was that dissimilar classes may have different discriminative features. 
The authors of \cite{zhou2011hierarchical} proposed the following orthogonal regularization: 
\begin{eqnarray}
\Omega(\vect {w_1},\cdots,\vect {w_m})=\sum_{ij}K_{ij}\left|\vect{w_{i}}^T\vect{w_{j}}\right|\\\nonumber
\end{eqnarray}
where $K_{ij}$ is the weight for $i$th and $j$th models. Minimizing this quantity will make the learned models orthogonal to each other. However, 
orthogonality does not necessarily imply different models will select different features, thus the learned models with such a constraint are 
not necessarily disjoint. For instance, the vector $\left[0.5,0.5\right]$ is orthogonal to the vector $\left[-0.5,0.5\right]$, however, it is 
obvious that the two vectors are not disjoint. \cite{hwanglearning} propose the following competition regularization term: 
 \begin{eqnarray}
 \Omega(\vect {w_1},\cdots,\vect {w_m})=&&\sum_{ij,i \neq j}\left\| \left|\vect{w_i}\right| + \left|\vect{w_j}\right| \right \|^2_2\\\nonumber
=&& \sum_{ij,i \neq j}(\left\| \vect{w_i} \right\|^2_2+\left\| \vect{w_j} \right\|^2_2+2\left|\vect{w_i}\right|^T\left|\vect{w_j}\right|)\\\nonumber
 \end{eqnarray}
This is very similar to the terms that control the model disjointness in problem~\ref{multi-1norm-svm}. 
The main difference is that here the importance of the $\ell^2_2$ norm is fixed with respect to that of
$\left|\vect{w_i}\right|^T\left|\vect{w_j}\right|$. In problem~\ref{multi-1norm-svm} we control the trade-off 
through the different $\lambda$ parameters, control which is crucial because we need to tune the appropriate model 
dissimilarity level.

We can easily extend the \mdsvm s learning problem to multi-class classification problems.
Given $m \times c$ models for a $c$-class classification problem, $\vect{w_{11}},\cdots,\vect{w_{mc}}$, 
the disjointness regularization term can be defined as:  
\begin{eqnarray}
\label{mc-disjoint-reg}
\Omega(\vect {w_{11}},\cdots,\vect {w_{mc}})=\sum_{ij, i\neq j}(\sum_l \left|\vect{w_{il}}\right|)^T(\sum_l \left|\vect{w_{jl}}\right|)\\\nonumber
\end{eqnarray}
Similar to the disjointness regularization in equation (\ref{disjoint-reg}), we want to force the 
different models to use different discriminative feature groups. However, the disjointness term now
regularizes model dissimilarity between the different $c$ model groups that correspond to the $c$
class classification problem.  
The maximizing prediction agreement of the multiple disjoint model 
learning for multi-class classification will depend on which methodology we will use to deal 
with the multi-class learning~\cite{hsu2002comparison,crammer2002algorithmic}. 

\subsection{Optimization}
\label{sec:opt}
Since the prediction agreement constraint in problem(\ref{multi-1norm-svm}) is not convex, the optimization 
problem (\ref{multi-1norm-svm}) is not convex either. However, if we fix all models except the $i$th one, then
the optimization problem of learning  $\vect{w_{i}},{b_{i}}$ becomes:
\begin{eqnarray}
\label{single-1norm-svm}
\min_{b_i,\vect{w_i},\vect \epsilon, \vect s}&&C+\sum_{k=1}^{n}\epsilon_{k}+\lambda_1\sum_{j=1,i\neq j}^m\sum_{k=1}^{n}s_{ijk}
+\lambda_2 \sum_{j, i\neq j}\left|\vect{w_i}\right|^T\left|\vect{w_j}\right|+\lambda_3 \left\| \vect{w_i} \right\|^2_2\\\nonumber
s.t.&&  y_k(b_i+\vect{w_i^T x_k}) \geq 1- \epsilon_{k}, \epsilon_{k} \geq 0, \forall {k}\\\nonumber
&&  (b_i+\vect{w_i^T x_k}) *(b_j+\vect{w_j^T x_k}) \geq -s_{ijk}, s_{ijk} \geq 0, \forall {j,k}\\\nonumber
\end{eqnarray}
where $C$ is a constant, the value of which is the sum of constant terms irrelevant with 
the $i$th model. Fortunately, this is a convex problem that is similar to the optimization 
problem of EN-SVM~\cite{ye2011efficient}. With some algebra, the objective function of 
(\ref{single-1norm-svm}) can be rewritten as:
\begin{eqnarray}
\label{weighted-1norm-svm}
C+\sum_{k=1}^{n}\epsilon_{k}+\lambda_1\sum_{j=1,i\neq j}^m\sum_{k=1}^{n}s_{ijk}+\sum_{l}(\lambda_2\sum_{j,j \neq i}\left|w_{jl}\right|)w_{il} 
+\lambda_3 \left\| \vect{w_i} \right\|^2_2\\\nonumber
\end{eqnarray}
Comparing optimization problem (\ref{weighted-1norm-svm}) to EN-SVM we see that the latter has different $\ell_1$ 
regularization weights for different features; for the $\ell$th feature the weight of the $\ell_1$ norm of $w_{il}$ 
is $ (\lambda_2\sum_{j,j \neq i}\left|w_{jl}\right|)$. From (\ref{weighted-1norm-svm}) we can see how the disjoint regularization works.
For example, if some models have already selected the $l$th feature, the $i$th model $\vect {w_i}$ will have a reduced probability of 
including the $l$th feature by increasing the weight of the $\ell_1$ norm for $w_{il}$. On the same time it will increase the probability 
to include other useful features which were not selected by the other models. 

Since the optimization problem (\ref{single-1norm-svm}) is convex, we will use 
the alternating convex optimization method to iteratively solve (\ref{multi-1norm-svm}). 
The details of the proposed algorithm are described in Algorithm (\ref{algo:MDSVM}). At each step, 
we learn only one of the $m$ models while the parameters of the rest are fixed. The convergence 
(possible to a local optima) of the alternating convex optimization 
method is guaranteed~\cite{bezdek2002some}. The main difficulty in optimizing the problem (\ref{single-1norm-svm}) is the non-differentiability of its objective function due to the $\ell_1$ regularization. As we will demonstrate our approach on microarray classification problem that typically has small number of high dimensional instances, we will solve (\ref{single-1norm-svm}) by the alternating direction method of multipliers (ADMM) following the work of~\cite{ye2011efficient}. However, for the large scale dataset with thousands of instances and features, the stochastic learning algorithm that exploiting the regularization structure, such as Regularized Dual Averaging method~\cite{xiao2010dual}, could be an alternative approach to optimizing the problem (\ref{single-1norm-svm}).


%
%
\begin{algorithm}[tb]
   \caption{\mdsvm s}
   \label{algo:MDSVM}
\begin{algorithmic}
   \STATE {\bfseries Input:} $\mathbf{X}$, $\mathbf{Y}$,$\lambda_1$,$\lambda_2$,$\lambda_3$,$m$
   \STATE {\bfseries Output:} $\vect{w_i}s$ and $b_is$ 
    \STATE initialize: $\vect{w^0_i}s=0,b^0_is=0$, and $i=1$
   \REPEAT
   \FOR {$j=1,\cdots,m$}
   \STATE Learning $(\vect{w_j}^{i},b^{i}_j)$ by solving the convex problem (\ref{single-1norm-svm})
   \ENDFOR
   \STATE $i:=i+1$
   \UNTIL{convergence}
\end{algorithmic}
\end{algorithm}

\section{Experiments}
\label{sec:exp}
In this section we will explore the behavior of our \mdsvm s algorithm on nine high dimensional biological datasets. 
The details of the datasets are given in Table (\ref{datasets}). All the datasets were preprocessed by standardizing 
the input features. The main goal of the experiments is to examine the degree to which the models produced by the 
\mdsvm s algorithm are \egd, i.e. the degree to which they satisfy the three properties of the \egd s given in 
definition~\ref{EGD}, namely the disjointness property, the high predictive agreement property, and the similar 
predictive power to a model learned on the complete set of relevant feature $S$.
To acquire the latter we use the EN-SVM algorithm, taking special care to avoid any information leakage between 
training and testing as we will see later in the full description of the experimental setup. 

To evaluate the degree to which the multiple models have a predictive performance which is comparable to that
of the single model learned over the $S$ set we will compare them against a standard single linear SVM model 
learned on $S$. 
\note[Removed]{
For comparison reasons we will also include a $\ell_1$-norm SVM~\cite{zhu20041} trained also on the $S$ feature set. }
In addition we also want to examine the information content of the \ifg\ and \cfs\ 
feature sets that are established as a result of the application of the \mdsvm s. To do so we will use the \mdsvm s 
to establish these feature sets and subsequently train over them a standard linear SVM.

For the EN-SVM we select values of the parameters that determine the importance of the $\ell_1$ and $\ell_2^2$ norms 
from the sets $\{0.1,1,10\}$ and $\{1,10,100\}$ respectively using a two-fold inner Cross-Validation (CV) on 
the training set. \note[Removed]{For the $\ell_1$-norm SVM, we select the value of the parameter that controls the importance of the
$\ell_1$ norm from the $\{0.1,0.5,1,5,10\}$ set also by two-fold inner CV.}
We set the value of the parameter of the $\ell_2^2$ norm of linear SVM that we use to estimate the predictive power of the different feature sets to the
value of the respective EN-SVM parameter.
The \mdsvm s algorithm has three hyper-parameters, problem (\ref{multi-1norm-svm}). To reduce 
the computational burden we set the value of $\lambda_1$ to one which is large enough 
to achieve high prediction agreement between models. We tune the remaining two parameters that 
control the model disjointness, i.e. $\lambda_2$, $\lambda_3$, as well as the number of models 
$m$ using inner 2-fold CV.  A large $\frac{\lambda_2}{\lambda_3}$ ratio corresponds to more dissimilar 
models.  We select $\lambda_3$ from $\{0.1,1,10,100\}$ and the value of $\lambda_2$ 
from $\{3,5,7,10\}*\lambda_3$. The number of models $m$ is selected from $\{1,2,3,4,5\}$. 

\mdsvm s needs to be trained on the set of relevant features $S$ which 
we establish through EN-SVM. Given a training and a testing set, $tr$ and $ts$ respectively, of some
fold we select the best parameter setting for EN-SVM, $\vect \lambda^{*}_{E}$, on $tr$ with two-fold inner CV. 
We now tune the parameters of \mdsvm s also by two-fold inner CV on $tr$ where in each fold the 
relevant feature set $S$ is given by the application of EN-SVM with the $\vect \lambda^{**}_{E}$ parameter 
setting that tuned on the $tr^*$ of this fold. Once we select the appropriate setting for the  \mdsvm s we reapply it on the $tr$ set
to produce the $m$ models which we then test on the $ts$ set.

We should note here that the objective function of \mdsvm s uses both the predictive performance and the model dissimilarity which are 
in an antagonistic relation, i.e. higher dissimilarity most often leads to lower predictive performance. This means 
that if we use only the classification error to guide the parameter selection for \mdsvm s in the inner CV most often 
we will arrive to configurations that have the smallest model disjointness. However we would still like to have a 
certain tolerance for model dissimilarity, since we want to get diverse models. In order to achieve that we use a 
trade-off between the classification error and the model dissimilarity to select the best parameter setting. More 
precisely the evaluation quantity that will drive the parameter selection for \mdsvm s is now: 
\begin{eqnarray}
\label{cv-criterion}
z=(1+Dis*\sigma\%)*P
\end{eqnarray}
where $P$ is the average accuracy of the learned models, and $Dis$ is the average pairwise model dissimilarity 
estimated by the inner 2-fold CV. We select the parameter setting that optimizes $z$. $\sigma$ controls the trade-off 
between accuracy and dissimilarity, with larger values of $\sigma$ favoring mode dissimilar models. As we are learning \egd s,
here we set $\sigma=2$ to make their predictive performance similar to that of the model learned on the $S$ feature set. 

The $\vect{w_i}$s and $b_i$s parameters of the $m$ models that will be learned by \mdsvm s are 
initialized to $\frac{1}{m}\vect w$ and $\frac{1}{m}b$, where $\vect w$ and $b$ are ones 
learned by EN-SVM. Since with \mdsvm s we learn $m$ models, we use as its predictive 
performance the average predictive performance of the $m$ models.  To estimate the predictive 
performance for each dataset we generated 10 random splits to training and testing. In each split, 80\% 
of the instances were used for training and the rest for testing. 
\note[Removed]{
The statistical significance of the differences were tested using t-test with a p-value of 0.05. }

\begin{table}[bt]
\begin{center}
\caption{Examined datasets.}
\label{datasets}
\vskip 0.15in
 \scalebox{0.8}{
\begin{tabular}{l||ccc||l||ccc} 
  Datasets                  & \# Sample& \# Feature &  \# Class&  Datasets                  & \# Sample& \# Feature &  \# Class \\ \hline \hline
   Lung                         & 39      &1971       &2 & Male vs. Female             & 134     & 1524      & 2  \\
   Breast$^1$                   &60       &1368       &2 &CNS                      & 60     & 7129      & 2  \\
   Breast$^2$                   &58       &3389       &2&Leukemia                    & 72      & 7129      & 2 \\
   Breast$^3$                   &49       &7129       &2&Ovarian                     & 253     & 771       & 2    \\
   Colon                       & 62      & 2000      & 2     & &&&           \\
\end{tabular}
}
\end{center}
\vskip -0.2in
\end{table}

\note[Removed]{
\begin{table}[bt!]
\begin{center}
\caption{Accuracy results. The superscripts $^{+-=}$ next to the accuracies of \mdsvm s indicate the result 
of the t-test of the comparison of each accuracy to those of $\ell_1$-norm SVM; they denote respectively a significant 
win, loss or no difference.}
\label{results}
\vskip 0.15in
 \scalebox{0.68}{
\begin{tabular}{l||c|c|c|c|cccc}
{\multirow{2}{*}{Datasets}} & {\multirow{2}{*}{$\ell_1$ norm SVM}}  & {\multirow{2}{*}{SVM-Union}}  & {\multirow{2}{*}{SVM-FSSFS}}  & {\multirow{2}{*}{SVM-NR}}  &\multicolumn{4}{|c}{\mdsvm s }  \\ \cline{6-9} 
           &&&& &  Accuracy  & \# Models & Agreement&Dis. Score \\ \hline 
Lung & 68.57$\pm$17.56 & 77.14$\pm$15.36 & 68.62$\pm$12.54 & 60.00$\pm$34.21 &$77.40$$\pm$12.58 & 3.50$\pm$1.27 & 0.92$\pm$0.10 &  0.65 $\pm$ 0.32 \\
Breast$^1$   & 68.33$\pm$17.48 & 75.83$\pm$9.17 & 71.94$\pm$10.10 & 65.83$\pm$13.86 &$73.53$$\pm$8.78 & 3.00$\pm$1.15 & 0.91$\pm$0.08 &  0.54 $\pm$ 0.34 \\
Breast$^2$     & 86.36$\pm$8.83 & 89.09$\pm$5.75 & 85.41$\pm$5.61 & 82.73$\pm$10.88 &$88.82$$\pm$6.53 & 3.50$\pm$1.35 & 0.97$\pm$0.03 &  0.41 $\pm$ 0.26 \\
Breast$^3$      & 56.67$\pm$12.23 & 54.44$\pm$13.30 & 55.39$\pm$10.20 & 28.89$\pm$33.62 &$58.24$$\pm$11.71 & 3.90$\pm$0.99 & 0.78$\pm$0.17 &  0.66 $\pm$ 0.45 \\
Colon & 85.83$\pm$11.15 & 85.00$\pm$10.24 & 83.56$\pm$9.79 & 75.00$\pm$28.60 &$85.56$$\pm$7.54 & 3.10$\pm$1.29 & 0.95$\pm$0.05 &  0.69 $\pm$ 0.32 \\
Male vs. Female & 83.85$\pm$5.68 & 90.00$\pm$3.72 & 83.31$\pm$4.86 & 63.08$\pm$19.38 &$86.11$$\pm$5.89 & 3.00$\pm$0.94 & 0.86$\pm$0.09 &  0.78 $\pm$ 0.20 \\
CNS& 63.33$\pm$14.80 & 75.00$\pm$13.61 & 71.39$\pm$10.23 & 46.67$\pm$32.68 &$72.08$$\pm$9.57 & 2.90$\pm$0.99 & 0.84$\pm$0.11 &  0.71 $\pm$ 0.34 \\
Leukemia & 97.14$\pm$3.69 & 97.14$\pm$3.69 & 96.51$\pm$4.20 & 55.00$\pm$47.62 &$97.29$$\pm$3.26 & 4.00$\pm$0.82 & 0.99$\pm$0.01 &  0.69 $\pm$ 0.32 \\
Ovarian & 99.00$\pm$1.41 & 98.20$\pm$1.99 & 96.44$\pm$2.21 & 96.60$\pm$2.99 &$98.15$$\pm$1.49 & 2.60$\pm$0.97 & 0.99$\pm$0.01 &  0.43 $\pm$ 0.17 \\
\hline
\end{tabular}
}
\end{center}
\vskip -0.2in
\end{table} }

\begin{table}[bt!]
\begin{center}
\caption{Accuracy results and statistics on the properties of the \mdsvm s models.}

\label{results}
\vskip 0.15in
 \scalebox{0.68}{
\begin{tabular}{l||c|c|c||cccc}
{\multirow{2}{*}{Datasets}}       & \multicolumn{3}{c}{SVM}  &\multicolumn{4}{|c}{\mdsvm s }  \\ \cline{2-8} 
                                  & $S$ & \cfs\ & \ifg      &  Accuracy  & \# Models & Agreement&Dis. Score \\ \hline 
Lung                              & 77.14$\pm$15.36 & 68.62$\pm$12.54 & 60.00$\pm$34.21 &$77.40$$\pm$12.58 & 3.50$\pm$1.27 & 0.92$\pm$0.10 &  0.65 $\pm$ 0.32 \\
Breast$^1$                        & 75.83$\pm$9.17 & 71.94$\pm$10.10 & 65.83$\pm$13.86 &$73.53$$\pm$8.78 & 3.00$\pm$1.15 & 0.91$\pm$0.08 &  0.54 $\pm$ 0.34 \\
Breast$^2$                        & 89.09$\pm$5.75 & 85.41$\pm$5.61 & 82.73$\pm$10.88 &$88.82$$\pm$6.53 & 3.50$\pm$1.35 & 0.97$\pm$0.03 &  0.41 $\pm$ 0.26 \\
Breast$^3$                        & 54.44$\pm$13.30 & 55.39$\pm$10.20 & 28.89$\pm$33.62 &$58.24$$\pm$11.71 & 3.90$\pm$0.99 & 0.78$\pm$0.17 &  0.66 $\pm$ 0.45 \\
Colon                             & 85.00$\pm$10.24 & 83.56$\pm$9.79 & 75.00$\pm$28.60 &$85.56$$\pm$7.54 & 3.10$\pm$1.29 & 0.95$\pm$0.05 &  0.69 $\pm$ 0.32 \\
Male vs. Female                   & 90.00$\pm$3.72 & 83.31$\pm$4.86 & 63.08$\pm$19.38 &$86.11$$\pm$5.89 & 3.00$\pm$0.94 & 0.86$\pm$0.09 &  0.78 $\pm$ 0.20 \\
CNS                               & 75.00$\pm$13.61 & 71.39$\pm$10.23 & 46.67$\pm$32.68 &$72.08$$\pm$9.57 & 2.90$\pm$0.99 & 0.84$\pm$0.11 &  0.71 $\pm$ 0.34 \\
Leukemia                          & 97.14$\pm$3.69 & 96.51$\pm$4.20 & 55.00$\pm$47.62 &$97.29$$\pm$3.26 & 4.00$\pm$0.82 & 0.99$\pm$0.01 &  0.69 $\pm$ 0.32 \\
Ovarian                           & 98.20$\pm$1.99 & 96.44$\pm$2.21 & 96.60$\pm$2.99 &$98.15$$\pm$1.49 & 2.60$\pm$0.97 & 0.99$\pm$0.01 &  0.43 $\pm$ 0.17 \\ \hline
\end{tabular}
}
\end{center}
\vskip -0.2in
\end{table}

\note[Removed]{
\begin{table}[bt!]
\begin{center}
\caption{Percentage of structured features. }
\label{length}
\vskip 0.15in
 \scalebox{0.7}{
\begin{tabular}{l||c|c|c|c}
Datasets&Union&FSSFS&NR& \mdsvm s \\ \hline 
Lung            & 98.61$\pm$2.87 \%  & 31.76$\pm$14.64 \%& 22.48$\pm$31.59 \% & 54.24$\pm$22.03 \%\\
Breast$^1$      & 99.96$\pm$0.13 \%  & 30.64$\pm$16.28 \%& 39.05$\pm$34.06 \% & 69.69$\pm$18.90 \%\\
Breast$^2$      & 100.00$\pm$0.00 \% & 36.61$\pm$14.98 \%& 39.96$\pm$26.86 \% & 76.57$\pm$13.94 \%\\
Breast$^3$      & 99.90$\pm$0.23 \%  & 25.11$\pm$17.72 \%& 29.21$\pm$45.89 \% & 54.32$\pm$31.88 \%\\
Colon           & 99.06$\pm$1.73 \%  & 36.04$\pm$15.91 \%& 22.02$\pm$33.75 \% & 58.06$\pm$18.57 \%\\
Male vs. Female & 99.92$\pm$0.10 \%  & 39.42$\pm$8.65 \%& 11.56$\pm$18.81 \% & 50.97$\pm$14.55 \%\\
CNS             & 99.90$\pm$0.20 \%  & 36.46$\pm$14.36 \%& 19.28$\pm$31.73 \% & 55.74$\pm$22.80 \%\\
Leukemia        & 99.28$\pm$0.70 \%  & 31.13$\pm$12.06 \%& 18.07$\pm$31.00 \% & 49.20$\pm$24.37 \%\\
Ovarian         & 99.49$\pm$0.69 \%  & 26.01$\pm$9.86 \%& 49.85$\pm$19.73 \% & 75.86$\pm$11.05 \%
\\\hline
\end{tabular}
}
\end{center}
\vskip -0.2in
\end{table} }

\begin{table}[bt!]
\begin{center}
\caption{Relative cardinalities of the features found/used in the \cfs, \ifg, and \mdsvm s, with respect to the cardinality of the $S$ feature set.}
\label{length}
\vskip 0.15in
 \scalebox{0.7}{
\begin{tabular}{l||c|c|c}
Datasets         & \cfs\ & \ifg & \mdsvm s \\ \hline 
Lung                                 & 31.76$\pm$14.64 \%& 22.48$\pm$31.59 \% & 54.24$\pm$22.03 \%\\
Breast$^1$                           & 30.64$\pm$16.28 \%& 39.05$\pm$34.06 \% & 69.69$\pm$18.90 \%\\
Breast$^2$                           & 36.61$\pm$14.98 \%& 39.96$\pm$26.86 \% & 76.57$\pm$13.94 \%\\
Breast$^3$                           & 25.11$\pm$17.72 \%& 29.21$\pm$45.89 \% & 54.32$\pm$31.88 \%\\
Colon                                & 36.04$\pm$15.91 \%& 22.02$\pm$33.75 \% & 58.06$\pm$18.57 \%\\
Male vs. Female                      & 39.42$\pm$8.65 \%& 11.56$\pm$18.81 \% & 50.97$\pm$14.55 \%\\
CNS                                  & 36.46$\pm$14.36 \%& 19.28$\pm$31.73 \% & 55.74$\pm$22.80 \%\\
Leukemia                             & 31.13$\pm$12.06 \%& 18.07$\pm$31.00 \% & 49.20$\pm$24.37 \%\\
Ovarian                              & 26.01$\pm$9.86 \%& 49.85$\pm$19.73 \% & 75.86$\pm$11.05 \%
\\\hline
\end{tabular}
}
\end{center}
\vskip -0.2in
\end{table}

We report the accuracy resuls in Table \ref{results}. The models of \mdsvm s have the same predictive
performance to that of the single SVM learned on the $S$ feature set in six of the nine datasets, while
for the three remaining, Breast$^1$, Male vs Female, and CNS, their performance is very similar.
\note[Removed]{
Compared to the performance of second baseline method, $\ell_1$ norm SVM, which learns a sparse 
model in the whole input feature space, the performance of MDSVMs is better in 
seven out of nine and almost the same on the rest Colon and Ovarian datasets.  }
For the single SVM models learned on the \ifg\ and \cfs\ feature sets, we see that their predictive 
performance is always worse than that of \mdsvm s.  This indicates that these feature sets contain predictive 
information that is complementary and should be used in the same model, as it is done by \mdsvm s, and not independently.
In terms of the prediction agreement of the models learned by \mdsvm s, we see that this is quite high, more than 85\%
with the exception of the Breast$^3$ dataset for which it is around 78\%, dataset for which the predictive performance
of all the methods was close to that of the default classifier. In terms of the number of features that the 
models of \mdsvm s use these range for the different datasets from 50\% to 76\% of the features of the $S$ feature 
set, Table~\ref{length}. The number of core features, i.e. the cardinality of the \ifg\ set, ranges over the 
different datasets from 11\% to almost 50\% of the features of the $S$ feature set. The average size of the \cfs\ feature sets
ranges from 25\% to 39\% of the features of $S$.

\section{Conclusion}
\label{sec:con}
Motivated by the limitation that current feature selection algorithms only provide a flat list of relevant features with no 
further information on the internal structure of that feature set we propose a new learning paradigm in which we try to uncover
the structure that underlines the set of relevant features of some learning problem. We do so by learning over this 
relevant feature set as many as possible equally good and dissimilar models, i.e. models that have a very 
high predictive power, high predictive agreement, and are defined over different subsets of the set of 
relevant features. These models structure the set of relevant features in a set of non-replaceable features, i.e. 
features that are always present over all the models, and to a set of functionally similar features sets which can 
be used interchangeably with no loss of predictive performance given the set of non-replaceable features.
This type  of feature structure can be extremely valuable for many application domains in which what drives
the analysis process is understanding the mechanisms that underly the learning dataset, a scenario that is 
typical in scientific knowledge discover problems. In order to achive this kind of feature structure we 
presented a novel multiple model learning algorithm which among other things controls the model dissimilarity,
in terms of the features that these models use, as well as the predictive model agreement. 
We demonstrate its ability to learn equally good and dissimilar models on a number of high 
dimensional microarray classification problems.  

There is considerable work that needs to be done in order to fully explore and exploit the possibilities that the new learning
paradigm that we propose opens as well as to understand better its behavior. We want to extend it to non-linear models, e.g. 
kernel methods, in order to discover non-linear distinct feature groups. We also want to constraint model disjointness in a more meaningful
manner typically using background knowledge on feature dependencies and interactions.

\bibliography{LMDM,alexandros}
\bibliographystyle{plain}
\end{document}